\RequirePackage{fix-cm}
\documentclass[smallextended]{svjour3}       
\smartqed  

\usepackage{graphicx}
\usepackage{amsmath,amsfonts}
\usepackage[export]{adjustbox}
\usepackage[linesnumbered,ruled]{algorithm2e}
\usepackage{natbib}
\usepackage{comment}
\usepackage{color}
\begin{document}

\title{On Support Vector Machines under a multiple-cost scenario
}


\author{Sandra Ben\'itez-Pe\~na        \and
        Rafael Blanquero \and
        Emilio Carrizosa \and
        Pepa Ram\'irez-Cobo 
}


\institute{S. Ben\'itez-Pe\~na \at
              IMUS, Instituto de Matem\'aticas de la Universidad de Sevilla. \\
              Departamento de Estad\'istica e Investigaci\'on Operativa, Universidad de Sevilla. Spain.\\
              Tel.: +34-955420861\\
              \email{sbenitez1@us.es}           
           \and
           R. Blanquero \at
              IMUS, Instituto de Matem\'aticas de la Universidad de Sevilla. \\
              Departamento de Estad\'istica e Investigaci\'on Operativa, Universidad de Sevilla. Spain.\\
              \and
              E. Carrizosa \at
              IMUS, Instituto de Matem\'aticas de la Universidad de Sevilla. \\
              Departamento de Estad\'istica e Investigaci\'on Operativa, Universidad de Sevilla. Spain.\\
              \and
              P. Ram\'irez-Cobo \at
              Departamento de Estad\'istica e Investigaci\'on Operativa, Universidad de C\'adiz. Spain.\\
              IMUS, Instituto de Matem\'aticas de la Universidad de Sevilla. \\
}

\date{Received: date / Accepted: date}

\maketitle

\begin{abstract}
Support Vector Machine (SVM) is a powerful tool in binary classification,  known to attain excellent misclassification rates. On the other hand, many realworld classification problems,  such as those found in medical diagnosis, churn or fraud prediction, involve misclassification costs  which may be different in the different classes. However, it may be hard for the user to provide precise values for such misclassification costs, whereas it may be much easier to identify acceptable  misclassification rates values. In this paper we propose a novel SVM model in which misclassification costs are considered by incorporating performance constraints in the problem formulation. Specifically, our aim is to seek the hyperplane with maximal margin yielding misclassification rates below given threshold values. Such maximal margin hyperplane is obtained by solving a quadratic convex problem with linear constraints and integer variables. The reported numerical experience shows that our model gives the user control on the misclassification rates in one class (possibly at the expense of an increase in misclassification rates for the other class) and is feasible in terms of running times.
\keywords{Constrained Classification \and Misclassification costs \and Mixed Integer Quadratic Programming \and Sensitivity/Specificity trade-off \and Support Vector Machines}
 \subclass{62P99 \and 90C11 \and 90C30}
\end{abstract}

\section{Introduction}
\label{introduction}

In supervised classification we are given a set $\Omega$ of individuals belonging to two or more different classes, and the final aim is to classify new objects whose class is unknown. Each object $i  \in \Omega$ can be represented by a pair $(x_i,y_i)$, where $x_i \in \mathbb{R}^m$ is the {attribute} vector and $y_i  \in \mathcal{C}$ is the class membership of object $i$.

\indent
A state-of-the-art method in supervised classification is the support vector machine (SVM), see \cite{vapkinnature,vapnik1998statistical,cristianini2000introduction,carrizosa2013supervised}.
{In its basic version, SVM addresses two-class problems}, i.e., $\mathcal{C}$ has two elements, say, $\mathcal{C} = \{ -1,+1\}$. The SVM aims at separating both classes by means of a linear classifier, $\omega^\top x + \beta = 0$, where $\omega$ is the so called \textit{score vector}.  We will assume throughout this paper that $\mathcal{C} = \{ -1,+1\}$ and refer the reader to e.g. \cite{allwein2000reducing} for the reduction of multiclass problems to this case.\\
The {linear} SVM classifier is obtained by solving the following convex quadratic programming (QP) formulation with linear constraints:

\begin{eqnarray}
\nonumber
  \min_{\omega, \beta, \xi} & \omega^\top \omega + {C_{+}\sum\limits_{i \in I : y_i = +1} \xi_i +  C_{-}\sum\limits_{i \in I: y_i = -1} \xi_i}& \\
 \nonumber
  \text{s.t.} & y_i(\omega^\top x_i + \beta) \geq 1 - \xi_i,& i \in I  \rlap{\footnotesize\quad \quad \quad \quad \quad \quad \quad\quad \quad \quad (SVM$(C_+,C_-)$)}\label{SVM$(C_+,C_-)$}\\
 \nonumber
   & \xi_i \geq 0& i \in I,\\
 \nonumber
\end{eqnarray}
\noindent where $I$ represents the set of training data, $\xi_i \geq 0$ are artificial variables which allow data points to be misclassified, {and $C_{+}, C_{-}>0$ are \textit{regularization parameters}} to be tuned that control the trade-off between margin minimization and misclassification errors. The case $C_+$ = $C_-$ is frequently considered in the literature, but the use of different regularization parameters for the different classes may allow for a better control of misclassification costs or unbalancedness. See e.g. \cite{lin2002support}.

Given an object $i$, it is classified in the positive or the negative class according to the sign of the so-called score function, $sign(\omega^\top x_i + \beta)$, while for the case $\omega^\top x_i + \beta = 0$, the object is classified randomly.

\noindent
{A mapping into a high-dimensional feature space may be considered {(\cite{Cortes1995})}, which allows us to transform this linear classification technique in a non-linear one using Mercer Theorem, \cite{mercer1909} and the so-called kernel trick, e.g. \cite{cristianini2000introduction}. {In this way we can address problems with a very large number of features, such as those encountered in personalized medicine (\citealt{doi101093biostatisticskxw018}).}}

\noindent
{Hence, the general formulation of SVM is}
\begin{eqnarray}
\nonumber
   {\max_{\mathbf{\lambda}}} & {-\dfrac{1}{2} \sum_{jk}\lambda_j\lambda_ky_jy_kK(x_j,x_k) + \sum_l \lambda_l}\\
 \nonumber
  \text{s.t.} &  {\sum_i \lambda_iy_i = 0}\\
 \nonumber
  & {0 \leq \lambda_i \leq C_{+},} & {i \in I : y_i = +1}\\
 \nonumber
 &  {0 \leq \lambda_i \leq C_{-},} & {i \in I : y_i = -1,}\\
 \nonumber
\end{eqnarray}
{where $K:\mathbb{R}^m\times\mathbb{R}^m \rightarrow \mathbb{R}$ is a kernel function and $\mathbf{\lambda}$ are the usual variables of the dual formulation of the SVM.}

\indent
As mentioned, the goal in supervised classification is to classify objects in the correct class. However, ignoring imbalance (either in the classes size, either in the misclassification cost structure) may have dramatic consequences in the classification task, see \cite{carrizosa2008multi,he2013imbalanced, prati2015class,Maldonado2017}. {For instance, for clinical databases, there are usually more observations of healthy populations than of the disease cases, and therefore smaller classification errors may be obtained for the first case. For example, for the well known \textit{Breast Cancer Wisconsin (Diagnostic) Data Set} from the UCI repository (\citealt{Lichman2013}), the number of sick cases (212) is smaller than the size of control cases (357). If a standard SVM is used for classifying the dataset, then {the estimated rates} (average values according to a 10-fold cross-validation approach), are depicted in Table~\ref{tab:wiscon}.
\begin{table}[htb]
\centering \small
\begin{tabular}{lll}
    \hline
         & Mean & Std \\  
    \hline
     \texttt{\% benign instances well classified} & 99\% & 1.7 \\ 
    \texttt{\% malign instances well classified} & {94.8\%} & {4.9}\\   
    \hline
 \end{tabular}
\caption{Performance of {standard} SVM {with Radial Function
Basis kernel} in \texttt{wisconsin}. Average values and standard deviations computed from 10 realizations.}
\label{tab:wiscon}
\end{table}
 Even though both rates are high, it might be of interest to increase the accuracy of predicting cancer, perhaps at the expense of deteriorating the classification rates in the other class. This problem will be addressed in this paper.

 In order to cope with imbalancedness, either in class size or structure of misclassification costs, different methods have been suggested, see \cite{bradford1998pruning,freitas2007cost,carrizosa2008multi,datta2015near}. Those methods are based on adding parameters or adapting the classifier construction, among others. For example, in \cite{carrizosa2008multi} a biobjective problem of simultaneous minimization of misclassification rate, via the maximization of the margin, and measurement costs, is formulated.

In this paper a new formulation of the SVM is presented, in such a way that the focus is not only on the minimization of the overall misclassification rate but also on the performance of the classifier in the two classes {(either jointly or separately)}. In order to do that, novel constraints are added to the SVM formulation. {The keystone of the new model is its ability to achieve a deeper control over misclassification in contrast to previously existing models.} The proposed methodology will be called Constrained Support Vector Machine (CSVM) and the resulting classification technique will be referred as CSVM classifier.

{The remainder of this paper is structured as follows.
In Section~\ref{sec:CSVM}, the CSVM is formulated as an optimization problem, and details concerning its feasibility are given. Section~\ref{sec:Results} aims to illustrate the performance of the new classifier. A description in depth about the experiments' design, real datasets to be tested as well as the obtained results will be given.
The paper ends with some concluding remarks and possible extensions in  Section~\ref{sec:Conc}. }


\section{Constrained Support Vector Machines}
\label{sec:CSVM}

{In this section the Constrained Support Vector Machine (CSVM) model is formulated as a Mixed Integer Nonlinear Programming (MINLP) problem (\citealt{Bonami2008186,Burer201297}), specifically in terms of a Mixed Integer Quadratic Programming (MIQP) problem.}

{This section is structured as follows. In Section~\ref{ssec:TM} some theoretical foundations that motivate the novel constraints are given. Then, in Section~\ref{subsection21} the formulation of the CSVM is presented. We will depart from the linear kernel case to later extend it to the general kernel case via the kernel trick. Finally, in Section~\ref{solveCSVM}, some issues about the CSVM formulation, as its feasibility, {shall be discussed.}}

\subsection{Theoretical Motivation}
\label{ssec:TM}

 As commented before, the aim of this work is to build a classifier so that {the user} may have control over the performance over the two classes. Specifically, given a set $\Omega =\{(x_i, y_i)\}_i$ {of data (a random sample of a vector $(X,Y)$ with unknown distribution)}, the target is to obtain a classifier such that $p \geq p_0$, where $p$ is the value of a performance measurement and $p_0$ is a threshold chosen by the user. {The} performance measure $p$ is chosen by the user at her convenience and may be selected among the following rates: true positive rate (TPR) or sensitivity, true negative rate (TNR) or specificity and accuracy (ACC), which are defined as follows:
\begin{eqnarray}\label{performancemeasures}
\nonumber \textrm{TPR}: & p=P({\omega}^\top X + \beta > 0 | Y= +1) &\\
\textrm{TNR}: & p=P({\omega}^\top X + \beta < 0 | Y= -1) &\\
\nonumber \textrm{ACC}: & p = P(Y({\omega}^\top X + \beta) > 0 ), &
\end{eqnarray}
see for example, \cite{bewick2004statistics}.

In this paper, for the sake of clarity, the positive class shall be identified with the class of interest to be controlled.
For instance, in cancer screening studies, cancer is labelled as positive class whereas absence of cancer is labelled as negative. Also, in credit-scoring applications the positive class will be the defaulting clients. More examples will be discussed in Section 3.

If the random variable {$Z$, defined as}
$$
Z =
\left\{
\begin{tabular}{cl}
1, & if an observation is well classified,\\
0, & otherwise, \\
\end{tabular}
\right.
$$
\noindent {is considered,} then, the values of $p$ as in (\ref{performancemeasures}) corresponding to the probability of correct classification can be rewritten as
\begin{eqnarray}\label{performancemeasures2}
\nonumber \textrm{TPR}: & {p}=E[Z| Y=+1]\\
\nonumber \textrm{TNR}: & {p}=E[Z| Y=-1]\\
\nonumber \textrm{ACC}: & {p}=E[Z]
\end{eqnarray}
and estimated from an independent and identically distributed (i.i.d.) sample $\{Z_i\}_{i \in S}$, by
\begin{eqnarray}\label{performancemeasures2}
\nonumber \textrm{TPR}: &  \hat{p} = \bar{Z}_+ = \frac{\sum\limits_{i \in S_+ }Z_i}{|S_+|}\\
\nonumber \textrm{TNR}: &  \hat{p} = \bar{Z}_- = \frac{\sum\limits_{i \in S_- }Z_i}{|S_-|}\\
\nonumber \textrm{ACC}: &  \hat{p} = \bar{Z} = \frac{\sum\limits_{i \in S}Z_i}{|S|},
\end{eqnarray}
{where $S_+$ and $S_-$ denote, respectively, the subsets $\{ i \in S : y_i =+1 \}$ and $\{ i \in S : y_i =-1 \}$.}

From a hypothesis testing viewpoint, our aim is to build a classifier such that, for a given sample, one can reject the null hypothesis in
$$
\left\{
\begin{array}{ll}
 H_0: & p \leq p_0  \\
 H_1: & p > p_0.
\end{array}
\right.
$$

Under the classic decision rule, $H_0$ is rejected if $\hat{p} \geq p_0^*$ assuming that $\alpha = P(\text{type I error})$. From Hoeffding Inequality (\citealt{1023072282952}),
{\begin{eqnarray}\label{HoefIneq1}
P(\hat{p} \geq  p + c) \leq \exp(-2nc^2).
\end{eqnarray}
As $\alpha = P(\text{type I error}) = P(\hat{p} \geq p_0^* | p = p_0)$, substituting $p$ by $p_0$ in (\ref{HoefIneq1}) yields
}
\begin{eqnarray}\label{HoefIneq}
P(\hat{p} < p_0 + c) \geq 1-\exp(-2nc^2) = {1-\alpha},
\end{eqnarray}
{where $p_0 + c= p_0^*$. T}herefore, we can take
\begin{eqnarray}\label{HoefIneq2}p_0^* = p_0 + \sqrt{\dfrac{\log \alpha}{-2n}}.
\end{eqnarray}
{Note that $n$ equals $|S_+|$, $|S_-|$ or $|S|$, respectively, when considering the TPR, the TNR or the accuracy.}

Here, the selection of the Hoeffding Inequality is motivated by its distribution-free character, but other options as the Binomial-Normal approximation could have been chosen instead.

\subsection{CSVM formulation}
\label{subsection21}

In this section, the CSVM formulation is presented. As it will be seen, the formulation includes novel performance constraints, which make the optimization problem a MIQP problem in terms of some integer variables.

{We assume to be given a dataset with known labels. From such set we identify the training set $I$, used to build the classifier, and the anchor set $J$, used to impose a lower bound on the classifier performance. These sets will be considered disjoint.}

{{With the purpose of building} the CSVM,} the performance constraints will be formulated in terms of binary variables $\{z_j\}_{j \in J}${, which are realizations of the variable $Z$ in Section~\ref{ssec:TM} and defined as:}
\indent
$$
z_j =
\left\{
\begin{tabular}{cl}
1, & if instance $j$ is {counted as} well classified \\
0, & otherwise. \\
\end{tabular}
\right.
$$

In order to formulate the CSVM, novel constraints are added to the standard soft-margin SVM formulation as follows:

\begin{eqnarray}
\nonumber
  \min_{\omega, \beta, \xi,z} &  \omega^\top \omega + {C_{+}\sum\limits_{i \in I : y_i = +1} \xi_i + C_{-}\sum\limits_{i \in I : y_i = -1} \xi_i}& \\
 \label{usSVM1}
  \textrm{s.t.} & y_i(\omega^\top x_i + \beta) \geq 1 - \xi_i,& i \in I \\
  \label{usSVM2}
   & \xi_i \geq 0& i \in I\\
 \label{zwellclas}
  &  y_j(\omega^\top x_j + \beta) \geq 1 - {M_1}(1-z_j),& j \in J \rlap{\footnotesize\quad \quad \quad \quad \quad \quad \quad\quad \quad \quad (CSVM$_0$)}\label{CSVM_0} \\
   \label{zbin}
   & z_j \in \{ 0,1\} & j \in J\\
   \label{zconst}
   & {\hat{p}_\ell \geq {p_{0}^*}_{\ell}}  & \ell \in L.
\end{eqnarray}

In the previous optimization problem, {(\ref{usSVM1}) and (\ref{usSVM2}) are the usual constraints in the SVM formulation.} Constraints (\ref{zwellclas}) ensure that observations $j \in J$ with $z_j = 1$ will be correctly classified, without imposing any restriction when $z_j=0$, provided that  {$M_1$} is big enough. 
A collection of requirements on the performance of the classifier over $J$ can be specified by means of (\ref{zconst})
. Also, $L$ is the set of indexes of the constraints that has the form of (\ref{zconst}). {These constraints can be modeled via the binary variables $z_j$, for instance:
\begin{eqnarray}\label{performancemeasures30}
\nonumber \textrm{TPR}: &  \sum\limits_{j\in {J_+}}z_j \geq p_0^* | {J_+}|\\
\nonumber \textrm{TNR}: &  \sum\limits_{j\in {J_-}}z_j \geq p_0^* |  {J_-} |\\
\nonumber \textrm{ACC}: &  \sum\limits_{j\in J}z_j \geq p_0^*|J|,
\end{eqnarray}
where $J_+$ and $J_-$ denote, respectively, the subsets $\{ i \in J : y_i =+1 \}$ and $\{ i \in J : y_i =-1 \}$.
}

\noindent
{As before, by} considering the (partial) dual problem of (CSVM$_0$) and the \emph{kernel trick}, the general formulation of the CSVM is obtained as follows (the intermediate steps can be found in Appendix A):
\begin{eqnarray}
\nonumber\min\limits_{{\lambda},{\mu}, \beta, {\xi},{z}}  &  \sum\limits_{s,s' \in I} \lambda_s y_s \lambda_{s'} y_{s'} K({x}_s,{x}_{s'}) +
\sum\limits_{t,t' \in J} \mu_t y_t \mu_{t'} y_{t'} K({x}_t,{x}_{t'}) \\
\nonumber & +
2 \sum\limits_{s \in I, t \in J} \lambda_s y_s \mu_t y_t K({x}_s,{x}_t) {+ C_{+} \sum\limits_{i \in  I : y_i = +1 } \xi_i + C_{-} \sum\limits_{i \in  I : y_i = -1 } \xi_i} \\
\nonumber\mbox{s.t.} & z_j \in \{0,1\} & j \in J  \\
\nonumber & {\hat{p}_\ell \geq {p_{0}^*}_{\ell}}  & \ell \in L\\
\nonumber & {y_i \left( \sum\limits_{s \in I} \lambda_s y_s K({x}_s,{x}_i)+ \sum\limits_{t \in J} \mu_t y_t K({x}_t,{x}_i) + \beta \right) \geq 1 - \xi_i}  & {i \in I}\\
\nonumber & { y_j\left( 	 \sum\limits_{s \in I} \lambda_s y_s K({x}_s,{x}_j)+ \sum\limits_{t \in J} \mu_t y_t K({x}_t,{x}_j)+ \beta \right) \geq 1-{M_1}(1-z_j) } & j \in J \rlap{\footnotesize\quad \quad   \quad (CSVM)}\\
\nonumber     &   \xi_i \geq 0 & i \in I \\
\nonumber     & {\sum\limits_{i \in I} \lambda_i y_i  + \sum\limits_{j \in J} \mu_j y_j = 0}\\
\nonumber &    0 \leq \lambda_i \leq {C_{+}}/2 & i \in I {: y_i = +1} \\
\nonumber &    0 \leq \lambda_i \leq {C_{-}}/2 & i \in I {: y_i = -1} \\
\nonumber & 0 \leq \mu_j \leq {M_2} z_j & j \in J.
\end{eqnarray}
Here $K:\mathbb{R}^m\times\mathbb{R}^m \rightarrow \mathbb{R}$ is {again} a kernel function, {$M_1$ and $M_2$ are big enough numbers}, {and $(\lambda,\mu)$ are the usual variables in the dual formulation of the SVM.}

\subsection{Solving the CSVM}
\label{solveCSVM}
In this section we give details about the complexity of our problem as formulated in (CSVM).
The problem belongs to the class of MIQP problems, and thus it can be addressed by standard mixed integer quadratic optimization solvers. In particular, the solver Gurobi (\citealt{gurobi}) and its Python language interface (\citealt{VanRossum2011IP2011927}) have been used in our numerical experiments. In contrast to the standard SVM formulation, which is a continuous quadratic problem, the CSVM is harder to solve due to the presence of binary variables. Hence, the optimal solution may not be found in a short period of time; however, as discussed in our numerical experience, good results are obtained when the problems are solved heuristically by imposing a short time limit to the solver.\\
Performance constraints (\ref{zconst}) may define an infeasible problem since the values of the {${p_0^*}_\ell$}  may be unattainable in practice. {Hence, the study of the feasibility of Problem~(CSVM) is an important issue.} {As an example, consider data composed by two different classes, each one represented respectively by black and white dots in the top picture in Figure~\ref{fig:puntos}.
\begin{figure}[h!]
  \centering
  \includegraphics[width=12.8cm]{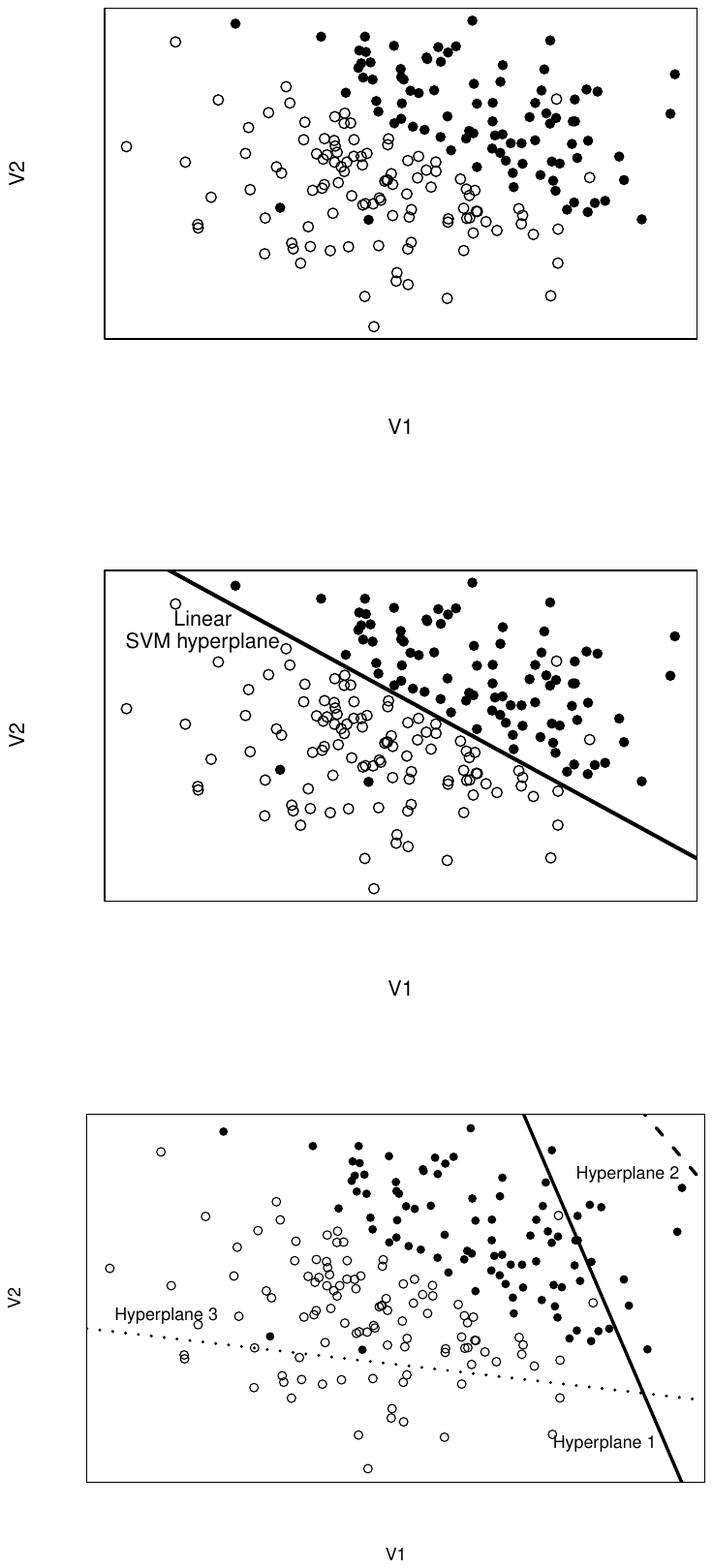}\caption{Study of feasiblity and unfeasibility of the CSVM.}\label{fig:puntos}
\end{figure}
If the optimization problem for the linear kernel SVM is solved, the resulting classifier is {a hyperplane that aims at separating both} classes and maximizes the margin. An approximate representation of the data and the classifier is shown in the middle panel in Figure~\ref{fig:puntos}.
If the aim is to correctly classify all the data corresponding to a given class, it is intuitively easy to see that this objective can be reached by moving the SVM hyperplane. In fact, it can be seen in the bottom picture in Figure~\ref{fig:puntos} how hyperplanes 1 and 2 classify correctly all white points, and hyperplane 3 classifies all the black dots in the correct class.  Among all those hyperplanes, the SVM selects the one which maximizes the margin. So, intuitively, it is evident that if just one constraint of performance is imposed in only one of the classes, the problem is always feasible. However, and using the data in Figure~\ref{fig:puntos} again, as well as the linear kernel SVM, it is clear that it is impossible to classify correctly all the instances at the same time; thus, the problem is then infeasible. }{However, there exist results, as Theorem 5 in \cite{burges1998tutorial}, that show that the class of Mercer kernels for which $K(x,x') \rightarrow 0$ as $\| x-x'\| \rightarrow \infty$, and for which $K(x,x)$ is $O(1)$, builds classifiers that get a total correct classification in all the classes in the training sample, without regard how arbitrarily the data have been chosen. Thus, if a kernel satisfies the previous conditions, then feasibility is guaranteed. In particular, Radial Function Basis (RBF) kernel meets these conditions. Therefore, to be on the safe side, if the performance thresholds imposed are not too low, they should refer only to one class misclassification rates (so that we can shift the variable $\beta$ to make the problem feasible) or to use a kernel, such as the RBF, known to have large VC dimension (\citealt{burges1998tutorial,cristianini2000introduction}), {defined as the {maximal training sample size for which perfect separation can always be enforced}}}.

\section{Computational results}
\label{sec:Results}

This section illustrates the performance of the novel method, the CSVM, in comparison with benchmark approaches. To do that, an assortment of datasets with different properties concerning size and unbalanceness shall be analyzed. Section \ref{subsec:Exper} describes the experiments to be carried out, while Section \ref{subsec:ParamSetting} details the choice of parameters. Section \ref{subsec:PerfEstim} is devoted to clarify different aspects of the cross-validation procedure for estimating the performance of the approach, and Section \ref{subsec:DataDesc} presents the datasets to be analyzed. Finally, Section \ref{subsec:Results} contains the obtained results and a deep discussion about them.

\subsection[Experiments]{{Description of the experiments}}\label{subsec:Exper}
{

{The objective of this paper, as has been stated before, is to build a classifier whose performance can be controlled by means of some constraints, as in Problem~(CSVM). As explained in Section~\ref{ssec:TM}, if we want a performance measurement $p$ to be greater than a value $p_0$ with a specified confidence $100(1-\alpha)\%$, we should use an estimator of $p$, $\hat{p}$, and impose it to be greater than $p_0^*=p_0+\sqrt{\dfrac{\log \alpha}{-2n}}$, {according to (\ref{HoefIneq2})}.}\\
{{Experiments whose} aim will be to increase the performance rate of interest in one class {will be performed}. However, as it will be shown, a damage may be produced in the other class. In particular, since the interest is to improve the classification in the positive class, the TPR will be the rate to be included in the novel constraints. Assume that an estimator of the TPR, TPR$_0$ is given. The aim will be to impose TPR $\geq$ TPR$_0$ $+ \delta_1$, were $\delta_1=0.025$, although other values can also be tested. {Therefore, our experiments will consist of:}
\begin{center}
 Impose TPR $\geq \min\left\{1,\text{TPR}_0+0.025\right\} {= p_0}$,
\end{center}}}
\noindent {which implies that, for $\alpha = 0.05$, the performance constraints in the optimization problem defining the novel CSVM are:
\begin{center}
$\widehat{\text{TPR}} \geq \min\left\{1,\text{TPR}_0+\sqrt{\dfrac{\log 0.05}{-2n}}+0.025\right\} {= p_0^*}$.
\end{center}
}

The novel CSVM will be compared with benchmark approaches. The first method to be compared with is the classic SVM where two different values of $C$ ($C_+$ and $C_-$) are used for each class. This approach shall be noted as SVM$(C_+,C_-)$ (see Section 1). The second benchmark method consists of moving the original hyperplane resulting from performing the standard SVM until the value $ p_0^*$ {obtained by Hoeffding Inequality} is achieved. This approach will be called from now on \textit{Sliding $\beta$ strategy}. {{}

\subsection{Parameters Setting}\label{subsec:ParamSetting}

 {One of the most popular kernels $K(x,x')$ in literature is the well-known RBF kernel (\citealt{cristianini2000introduction,hastie01statisticallearning,hsu2003practical,Smola2004,Horn2016}), given by
$$
K(x,x')= \exp\left( -\gamma \| x-x'\|^2\right),
$$
where $\gamma > 0$ is a parameter to be tuned. This will be the kernel chosen for implementing the CSVM, although the method is valid for an arbitrary kernel.}

 The time limit for the solver {was set equal to 300 seconds. In addition, the $M_1$ and $M_2$ values in Problem~(CSVM) were set both equal to 100.} {The choice of these values is motivated as follows. First, for the sake of computational tractability, the time limit should not be too high, but high enough so that the optimizer is able to solve the problem or at least to provide good feasible solutions. In our experiments, the choice of a time limit equal to 300s gave a good balance between the computational cost and the quality of the solutions. In the case of the values of $M_1$ and $M_2$, if small values are chosen, there may be many discarded hyperplanes, including the optimal one. However, if $M_1$ and $M_2$ are too big, it might cause {computational difficulties} (\citealt{doi101287inte251}) because of numerical instabilities and large gaps in the continuous relaxation, making the branch and bound too slow. A compromise solution is obtained by considering $M_1 = M_2=100$ in our problems. Setting $M_1$ and $M_2$ equal to $100$, not a huge number, may indeed exclude the optimal solution of the original problem. This is not a big issue, since the original problem is nothing but a surrogate of our real aim, namely, classifying correctly forthcoming individuals. On the other hand, this constraint may also be seen as a regularization constraint, since it forces the variables involved to take relatively small values, as already happens with the variables $\lambda_i$, already force to be below $C/2$. In other words, though at the expense of excluding the optimal solution of the proxy optimization problem, setting a not too large value for $M_1$ and $M_2$ can be seen as an extra regularization, thus preventing overfit.}}

 {Note that an alternative formulation, avoiding big $M$ constraints is obtained by using the Specially Ordered Sets of Type 1 (SOS1) (\cite{bertsimas2005optimization}, see also \cite{silva2017optimization} and \cite{bertsimas2016best} for some examples of SOS1). However, we prefer to maintain the big $M_1$ and $M_2$ for two reasons. On the one hand, the use of SOS1 would involve quadratic constraints, which would make the problem even more difficult to solve. For example, constraint $0 \leq \mu_t \leq M_2z_t$ would become $(\mu_t,1-z_t)$:SOS1 and $0 \leq \mu_t$. This is equivalent to $\mu_t(1-z_t)=0$ and $0 \leq \mu_t$, which includes, as we can see, a non-convex quadratic constraint. On the other hand, not every solver has implemented the SOS1 method or is capable to solve quadratic mixed integer problems with non-convex quadratic constraints. In addition, even if it can manage SOS1-type constraints, it might perform the conversion to the problem with a big $M$ automatically, and thus we would be again with big $M$ constraints, now controlled by the solver and not by ourselves.
 }

\subsection{Performance estimation}\label{subsec:PerfEstim}

{The estimation of the performance of the novel CSVM is based on a $K$-fold cross validation (CV) as follows, see \cite{kohavi1995study}. Generally, K=10, but for those datasets with more than 1000 samples, $K=5$ so that the running times are lower. Note that, apart from tuning $\gamma$, the regularization parameters $C_+$ and $C_-$ introduced in Section~\ref{introduction} also need to be tuned. {In order to make the {CSVM} procedure quicker, our experiments are based on choosing {$C_+ = C/|I_+|$ and $C_- = C/|I_-|$, so only one parameter $C$ shall be tuned {for the CSVM}, but not for the SVM$(C_+,C_-)$, in which both $C_+$ and $C_-$ are tuned independently}. As it will be seen later, this is not a crucial issue. Hence,} for a given pair of parameters $(C,\gamma)$, the process consists mainly on solving a standard SVM using all the instances ($I \cup J$), and collect the values of $\lambda$ (from the dual formulation of the SVM) as well as the value of $\beta$. Once the SVM is solved, and with the purpose of providing an initial solution for the CSVM, the value of $\beta$ is slightly changed (maintaining the values of $\lambda$'s fixed) until the desired number of instances well classified is reached. Then, the values of $\beta$ and $\lambda$'s obtained are set as initial solutions for CSVM. In addition, depending on whether each instance in $J$ is well classified or not, we set their values of $z$ as $0$ or $1$ as initial values for the CSVM.}

{We should make the selection of the best pair $(C,\gamma)$ in each of the previous folds. In order to do that, a 10-fold CV (5-fold CV for datasets with more of 1000 samples, in order to reduce the running times) as before is made for each pair in a grid given by the {121} different combinations of
{$C=2^{ (-5:5)}$ and $\gamma = 2^{ (-5:5)}$} ({$C_+=2^{ (-5:5)}$, $C_-=2^{ (-5:5)}$ and $\gamma = 2^{ (-5:5)}$} for SVM$(C_+,C_-)$). {The general criterion used to select the best pair of parameters is the accuracy. However, in cases where the datasets are severely unbalanced in the classes size (when one of the classes has a weight less than a 30\% of the total size), {the G-mean (\citealt{Tang2009SMH16567021656726}){, which is defined as $\sqrt{TPR \times TNR}$,} is used to perform the parameter tuning instead.}}} {Finally, the average values of TPR and TNR obtained in the first CV, in addition to their standard deviations, are calculated.}

{For a better understanding, the previous algorithm is summarized in Algorithm~\ref{algFS2}.}

{\SetAlgoNoLine
\begin{algorithm}
    \SetKwInOut{Input}{Input}
    \SetKwInOut{Output}{Output}
      Split data ($D$) into ``\textit{folds}'' subsets, $ D=\{D_1,\ldots, D_{folds}\}.$ \\
    \For{kf = 1,$\ldots$,folds}{
      Set $Validation = D_{kf}$ and $I \cup J = D - \{ D_{kf} \}$.\\
      \For{each pair $(C,\gamma)$ in grid $(\{2^{(-5:5)}\},\{2^{(-5:5)}\})$}{
      Split $ D - \{ D_{kf} \} =D^*$ into ``\textit{folds2}'' subsets, $ D^*=\{ D^*_1,\ldots, D^*_{folds2}\}.$ \\
          \For{kf2 = 1,$\ldots$, folds2}{
          Set $Validation^* = D^*_{kf2}$ and $I^* \cup J^*  = D^* - \{ D_{kf2}\}$.\\
          Run standard SVM over $I^* \cup J^*$.\\
          Move $\beta$ of SVM until the instances are correctly classified.\\
          Run problem CSVM over $I^*$, $J^*$ with initial solutions from before.\\
          Validate over $Validation^*$, getting the accuracy ($ACC[kf2]$).\\
          }
          Calculate the average accuracy $(\sum_{kf2} ACC[kf2])/folds2=\overline{ACC}$.  \\
          \If{$\overline{ACC} \geq bestACC$}{
                Set $bestACC$ = $\overline{ACC}$,  $best\gamma$ = $\gamma$ and $bestC$ = $C$.\\
          }
      }
     Run standard SVM over $I \cup J$ with the parameters $best\gamma$ and $bestC$.\\
     Move  $\beta$ of SVM until the instances are correctly classified.\\
     Run problem CSVM  over $I$, $J$ with initial solutions from the previous step.\\
     Validate over $Validation$, getting the correct classification probabilities ($TPR[kf]$, $TNR[kf]$).
    }
    Calculate the average values for $TPR$ and $TNR$.

    \caption{Pseudocode for CSVM}\label{algFS2}
\end{algorithm}
}

{Finally, we want to clarify that for our experiments we have selected $I$ as the first half of $ I\cup J$ and $J$ as the second one.}

\subsection[Datasets]{Data description}\label{subsec:DataDesc}

The performance, in terms of correct classification probabilities and accuracy, is illustrated using {6} real-life datasets from the UCI and {Keel repositories} (\citealt{Lichman2013} and \citealt{alcala2009keel}). {In particular, the datasets are \texttt{australian} (Statlog (Australian Credit Approval) Data Set), \texttt{votes} (Congressional Voting Records Data Set), \texttt{wisconsin} (Breast Cancer Wisconsin (Diagnostic) Data Set), \texttt{german} (Statlog (German Credit Data) Data Set), {\texttt{pageBlocks} (Page Blocks Classification (Imbalanced: 0) data set) and \texttt{biodeg} (QSAR biodegradation Data Set)}.}

{Details concerning the distribution of the classes in the considered datasets are provided by Table~\ref{tab:data}.
\begin{table}[h!]

\centering \small

\begin{tabular}{llllllll}

     \hline
     Name           & $V$ & $|\Omega|$   &  {$|\Omega_+|$} (\%)  \\
     \hline
    \texttt{australian}& 14 & 690     &  307 (44.5\%) \\
    \texttt{votes}     & 16 & 435     &  267 (61.4 \%) \\
    \texttt{wisconsin} & {30} & 569     &  212 (37.3 \%)   \\
    \texttt{german}    & {45} & 1000    &   300 (30\%)   \\
    \texttt{pageBlocks} & {10} &  5472   & 558  (10.2\%)   \\
    \texttt{biodeg}    & {41} &  1055  &  356 (33.7\%)   \\
    \hline

 \end{tabular}

\caption{Details concerning the implementation of the CSVM for the considered datasets.}

\label{tab:data}

\end{table}
{The first two columns give the name and number of {attributes} for each set. The values $|\Omega|$ and $|\Omega_+|$ represent, respectively, the size for each dataset and the number of positive instances in $\Omega$. Finally, the percentage of positive instances is compiled in the last column.}

{Note that prior to running the different experiments, data have been standardized, that is to say, each {attribute} has zero mean and unit variance.}

{As a remark, we want to express that for the two biggest datasets (those that have more than 1000 samples), an alternative is proposed in order to reduce the computational times. First, to train the classifier, instead of using the training samples, we have built clusters of training points of the same class via the k-means method. The number of clusters was selected so that the proportion of original positive and negative instances was maintained. Also, we took into consideration the number of instances per cluster to train the SVM. In the validation sample, we kept the instances as they were originally.}

\subsection[Results]{Results}\label{subsec:Results}

In this section we illustrate the performance of the CSVM in comparison with the classic SVM, the SVM$(C_+,C_-)$ and the \textit{Sliding $\beta$ strategy}. As previously commented, the purpose will be to increase the TPR.} {Note that, even though from Section~\ref{solveCSVM} the CSVM problem is always feasible using the training sample, it may happen that the desired performance is not achieved in the validation sample.}\\

Table  \ref{tab:tpr}
\begin{table}[h!]
\centering \small

\begin{tabular}{llllllllllllll}

    \hline
     Name & &  SVM & SVM$(C_+,C_-)$ &  Sliding $\beta$  &   CSVM \\    
    \hline
        &  & Mean  & Mean  & Mean (Target) & Mean (Target)\\
        &  & (Std) & (Std) &    (Std)      & (Std) \\
    \hline

    \texttt{australian} & \texttt{TPR} &\textbf{0.83}  & {\textbf{0.806}}  & \textbf{0.821} (0.855) &  \textbf{0.903} (0.855)  \\
    &    &   (0.071) & (0.093) & (0.073) & (0.05)\\
                   &{\texttt{TNR}}  & 0.863  & {0.878}  & 0.855   & 0.772 \\
                   &     &   (0.079) & (0.088) & (0.068) & (0.081)\\
                   \hline
    \texttt{votes} & {\texttt{TPR}}  & \textbf{0.963} &  \textbf{0.945} & \textbf{0.971} (0.988) & \textbf{0.978} (0.988)\\ 
    &   & (0.04) & (0.042) & (0.037)  & (0.026)\\
                   &\texttt{TNR}   & {0.951}  & 0.941 & 0.91  & 0.922 \\ 
                   &   & (0.031) & (0.037) & (0.063) & (0.04)\\
                   \hline
    \texttt{wisconsin} & \texttt{TPR}  & \textbf{0.948} &  \textbf{0.962}  & \textbf{0.989} ({0.973})  & {\textbf{0.965} ({0.973})} \\
    & & (0.049) & (0.027) & (0.017) & (0.037)\\
    &{\texttt{TNR}} & 0.99  &  0.931 & 0.953  & {0.945 } \\ 
    & & (0.017) & (0.07) & (0.045) & (0.045)\\
                       \hline
   \texttt{german}& \texttt{TPR} & \textbf{0.464}  &  {\textbf{0.89}}  & \textbf{0.043} (0.65) & {\textbf{0.671} (0.65)} \\
    & & (0.103) & (0.08) & (0.023) & (0.164)\\
                   &{\texttt{TNR}}        & 0.847  & {0.407} &  0.996 & {0.668}\\
                   & & (0.031) &(0.069) & (0.009) & (0.111) \\
    \hline
    \texttt{pageBlocks}& \texttt{TPR} & \textbf{0.807} & \textbf{0.557} & \textbf{0.819} (0.832) & \textbf{0.859} (0.832)  \\
    & & (0.03) & (0.361) & (0.981) & (0.045)\\
                       & \texttt{TNR} & 0.988 & \textbf{0.901} &  0.981 & 0.965  \\
                       & & (0.004) & (0.088) & (0.006) & (0.012)\\
                           \hline
    \texttt{biodeg}& \texttt{TPR} & \textbf{0.783}  & \textbf{{0.793}} & \textbf{0.797} (0.808) & {\textbf{0.852} (0.808)} \\
    & & (0.084) & (0.083) & (0.095) & (0.057)\\
                       & \texttt{TNR} & 0.909 & {0.839} &  0.891  & {0.833}  \\
                       & & (0.032) & (0.037) & (0.037) & (0.05)\\
                 \hline

 \end{tabular}

\caption{Results under the SVM, SVM$(C_+,C_-)$, the \textit{Sliding $\beta$ strategy} and the novel CSVM. Target rate: TPR}

\label{tab:tpr}

\end{table}
reports the average rates (and under them, and in parenthesis, their standard deviations) obtained under the SVM, SVM$(C_+,C_-)$, \textit{Sliding $\beta$ strategy} and CSVM, for the experiment described in Section \ref{subsec:Exper}, that is, when $\widehat{\textrm{TPR}}\geq \min\left\{1, \textrm{TPR}_0+\sqrt{\dfrac{\log 0.05}{-2n}}+0.025\right\}$ is imposed. Also, the target values (in parenthesis in the third and forth columns) to be achieved for the TPR  are shown.

Some comments arise from the table. In the case of \texttt{australian}, we trivially considered ``$+$'' as the positive class and ``$-$'' as the negative. The SVM$(C_+,C_-)$ slightly improves the TNR when it is compared with the standard SVM, but yields a worse value in the TPR, which is the rate to be improved. When the \textit{Sliding $\beta$ strategy} is used, although a target value of 0.855 is imposed, even a lower value that the one got with the SVM is obtained, with a lower TNR value also. On the other hand, when the CSVM is used instead, the increase is not only of 0.025 points but of 0.073, obviously at the expense of the other class. Hence, the best TPR is obtained for the CSVM. \newline
   We shall analyze next the results for \texttt{votes}, which has two classes: ``democrat'' and ``republican''. Since in principle there is no interest in a better classification of one of the classes, the majority class (``democrat'') will be identified as the positive class. From the table it can be seen how the results under the SVM$(C_+,C_-)$ are poorer than under the classic SVM. If the \textit{Sliding $\beta$ strategy} is used instead, an increase in the TPR is obtained but the rate
does not achieve the target value. Even though the CSVM does not achieve the target value in the validation set, here again, this novel approach achieves the best TPR. \newline
   Concerning \texttt{wisconsin} dataset, it has two classes: ``malignant'' and ``benign''. Here, we consider as positive the ``malignant'' class, which is clearly the class of interest. The results for the SVM$(C_+,C_-)$ are better than those obtained under the SVM, but it does not achieve the target value. When the \textit{Sliding $\beta$ strategy} is used, the target value for the TPR is achieved, while reducing the value for the TNR with respect to the SVM. Then, when we use the CSVM, the TPR is a bit higher than when the SVM$(C_+,C_-)$ is used, but lower than the one obtained for the \textit{Sliding $\beta$ strategy}. The same happens for the TNR. In this case, the method that performs the best is the \textit{Sliding $\beta$ strategy}.\newline
   Next, we shall analyze \texttt{german} dataset, which is composed by two classes: good and bad credit risk. The class of interest and hence the positive one, is ``bad credit risk''. {Here, the SVM$(C_+,C_-)$ improves in a significant way the estimation of the TPR in comparison to the classic SVM; however, this is achieved at the expense of worsening the TNR. The \textit{Sliding $\beta$ strategy} performs very poorly in the case of the TPR but provides in contrast a very high TNR. The CSVM gets the most balanced result: the TPR exceeds the target values, and at the same time, the TNR is not notably affected.  \newline
   We next describe the results obtained for the \texttt{pageBlocks} dataset which, as it has been previously commented, is a strongly unbalanced dataset with a dimension higher than in the previous cases. The two classes for this dataset are ``text'' and ``graphic'' areas. In addition, the ``graphic'' areas instances are less frequent (10.2 \%). Assume that for this problem the interest is in distinguishing the ``graphic'' areas from the ``text'' areas, therefore, the class to be controlled will be the ``graphic'' one. The results {show how the SVM$(C_+,C_-)$ obtains the opposite effect than the pursued. Both the TPR and TNR are lower than when the classic SVM is used. In the case of using the \textit{Sliding $\beta$ strategy}, the TPR is increased but it does not reach the imposed target. On the other hand, the TNR is slightly reduced. For the CSVM, the target value in the TPR is reached, resulting in a small decrease in the TNR.} \newline
   Finally, we present the results for \texttt{biodeg}, with two classes: ``ready biodegradable'' and ``not ready biodegradable''. Originally, \cite{doi101021ci4000213}, classification models were used to discriminate ``ready biodegradable'' from ``not ready biodegradable'', being ``ready biodegradable'' considered as the positive class. Here again, SVM$(C_+,C_-)$ improves the TPR with respect to the classic SVM. The \textit{Sliding $\beta$ strategy} outperforms the SVM$(C_+,C_-)$ but only the CSVM obtains an estimated TPR larger than the imposed lower bound. Note that, in contrast, the TNR under the CSVM is slightly lower than the values under the benchmark approaches.

Overall, the target value is almost always achieved when the CSVM is used. In the cases this does not occur, we obtain a close value. However, although initially one may think that good results will be obtained for the \textit{Sliding $\beta$ strategy}, such naive procedure does not achieve the target value so frequently. The same occurs with the SVM$(C_+,C_-)$. Hence, we can conclude that the method that provides more control on the performance measures is the CSVM, which highlights the novelty of our proposal.

\section{Conclusions}
\label{sec:Conc}

In this paper, we propose a new supervised learning SVM-based method, the CSVM, with the purpose of controlling a specific performance measure. Such classifier is built via a reformulation of the classic SVM, where novel constraints including integer variables are added. The final optimization problem is a MIQP problem, which can be solved using standard solvers as Gurobi or CPLEX. {In order to guarantee that the performance rate is lower bounded by a fixed constant with a high confidence, some theoretical foundations are provided.} The applicability of this cost-sensitive SVM has been demonstrated by numerical experiments on benchmark data sets.

We conclude that it is possible to control the classification rates in one class, possibly, but not necessarily, at the expense of the performance on the other class. This highly contrasts with the naive approach in which, once the SVM is solved, its intercept is moved to enhance the positive rates in one class, necessarily deteriorating the performance in the other class. The results presented confirm the power of our approach.

Although, for simplicity, all numerical results are presented just adding one performance constraint, one constraint per class, as well as an overall accuracy, may be added in our approach. {Also for simplicity, we addressed here two-ways data matrices and two-class problems; however, this approach could be extended to the case when using more complex data as multi-class or \textit{multi-way} arrays (\citealt{101093biostatisticskxw057}), which are very common in biomedical research. {On the other hand, an alternative perspective for addressing the SVM regularization is to consider different norms (\citealt{Yao2014}).}}

{Finally, another possible extension is to perform a feature selection which uses the proposed constraints in order to control the misclassification costs, see \cite{BENITEZPENA2018}. Such a process is an essential step in tasks such as high-dimensional microarray classification problems (\citealt{doi101093biostatisticskxq023}).}

\section*{Acknowledgements}
This research is supported by Fundaci\'on BBVA, and by projects FQM329 and P11-FQM-7603 (Junta de Andaluc\'{\i}a, Spain) and  MTM2015-65915-R (Ministerio de Econom\'{\i}a y Competitividad, Spain). The last three are cofunded with EU ERD Funds. The authors are thankful for such support.

\section*{Appendix A: Derivation of the CSVM}

In this section, the detailed steps to build the CSVM formulation are shown. For that, suppose that we are given {the mixed-integer quadratic} model
 \begin{eqnarray}
\nonumber
  \min_{\omega, \beta, \xi,z} &  \omega^\top \omega + {C_+\sum\limits_{i \in I : y_i =+1} \xi_i  + C_-\sum\limits_{i \in I : y_i =-1} \xi_i } & \\
 \nonumber
  \textrm{s.t.} & y_i(\omega^\top x_i + \beta) \geq 1 - \xi_i,& i \in I \\
  \nonumber
   & \xi_i \geq 0& i \in I\\
 \nonumber
  &  y_j(\omega^\top x_j + \beta) \geq 1 - M_1(1-z_j),& j \in J \\
   \nonumber
   & z_j \in \{ 0,1\} & j \in J\\
   \nonumber
   & {\hat{p}_\ell \geq {p_{0}^*}_{\ell}}  & \ell \in L.
\end{eqnarray}

Hence, the problem above can be rewritten as
\[
\begin{array}{lllllll}
\min_{{z}}   & & &  {\min_{{\omega},\beta,{\xi}} }& {{\omega}^\top {\omega}  + {C_+\sum\limits_{i \in I : y_i =+1} \xi_i  + C_-\sum\limits_{i \in I : y_i =-1} \xi_i } }\\
\mbox{s.t.} & z_j \in \{0,1\} & j \in J  &  {\mbox{s.t.}} & {y_i \left( {\omega}^\top {x}_i  + \beta \right) \geq 1 - \xi_i} & {i \in I}\\
 & {\hat{p}_\ell \geq {p_{0}^*}_{\ell}}  & \ell \in L &   &y_j\left(\omega^\top x_j + \beta\right) \geq 1 - M_1(1-z_j),& j \in J \\
& &  & &  { \xi_i \geq 0} & {i \in I }.
\end{array}
\]

{We first develop the expression of the dual for the linear case and then we show how the kernel trick applies.} As a previous step we should consider the variables $z$ as fixed. Hence, having those variables fixed, the inner problem is rewritten as:
    \[
\begin{array}{llll}
  {\min_{{\omega},\beta,{\xi}} }& {{\omega}^\top {\omega}  + {C_+\sum\limits_{i \in I : y_i =+1} \xi_i  + C_-\sum\limits_{i \in I : y_i =-1} \xi_i } }\\
  {\mbox{s.t.}} & {y_i \left( {\omega}^\top {x}_i  + \beta \right) \geq 1 - \xi_i} & {i \in I}\\
    &y_j\left(\omega^\top x_j + \beta\right) \geq 1,& j \in J  : z_j =1 \\
    &y_j\left(\omega^\top x_j + \beta\right) \geq 1 - M_1,& j \in J : z_j = 0\\
 &  { \xi_i \geq 0} & {i \in I }.
\end{array}
\]

{As $M_1$ is a large number, the fourth constraints always result feasible, so they can be removed. Also, we can denote $\{j \in J  : z_j =1\}$ by $J(z)$, obtaining
    \[
\begin{array}{llll}
  {\min_{{\omega},\beta,{\xi}} }& {{\omega}^\top {\omega}  + {C_+\sum\limits_{i \in I : y_i =+1} \xi_i  + C_-\sum\limits_{i \in I : y_i =-1} \xi_i } }\\
  {\mbox{s.t.}} & {y_i \left( {\omega}^\top {x}_i  + \beta \right) \geq 1 - \xi_i} & {i \in I}\\
    &y_j\left(\omega^\top x_j + \beta\right) \geq 1,& j \in J(z) \\
 &  { \xi_i \geq 0} & {i \in I }.
\end{array}
\]}

{Hence, we can build the Lagrangian
$$
\begin{array}{ll}
  \mathcal{L}(\omega,\beta,\xi) = & {{\omega}^\top {\omega}  + {C_+\sum\limits_{i \in I : y_i =+1} \xi_i  + C_-\sum\limits_{i \in I : y_i =-1} \xi_i }} -  \sum\limits_{s\in I} \lambda_s (y_s(\omega^\top x_s+\beta) -1 + \xi_s) - \\
   & - \sum\limits_{t\in J(z)} \mu_t (y_t(\omega^\top x_t+\beta) -1) - \sum_{i' \in I} \delta_{i'} \xi_{i'}
\end{array}
$$
The KKT conditions are, therefore
\[
\begin{array}{llllll}
\dfrac{\partial \mathcal{L}}{\partial \omega} = 0 & \Rightarrow &  {\omega} & = & \sum\limits_{s \in I} (\lambda_s/2) y_s {x}_s+ \sum\limits_{t \in J(z)} (\mu_t/2) y_t {x}_t \\
\dfrac{\partial \mathcal{L}}{\partial \beta} = 0 & \Rightarrow & 0 & = & \sum\limits_{s \in I} \lambda_s y_s  + \sum\limits_{t \in J(z)} \mu_t y_t \\
\dfrac{\partial \mathcal{L}}{\partial \xi_i} = 0 & \Rightarrow & 0 & = & -\lambda_i -\delta_i + C_+ & i \in I:y_i =+1\\
\dfrac{\partial \mathcal{L}}{\partial \xi_i} = 0 & \Rightarrow & 0 & = & -\lambda_i -\delta_i + C_- & i \in I:y_i =-1\\
 & & 0 & \leq & \lambda_{i} & i \in I\\
 & & 0 & \leq & \mu_t & t \in J(z)\\
 & & 0 & \leq & \delta_{i} & i \in I
\end{array}
\]}

{Note that we can replace, without loss of generality, $\lambda_s/2$, $\mu_t/2$ by $\lambda_s$ and $\mu_t$, respectively. Then, in the condition  ${\partial \mathcal{L}}/{\partial \beta} = 0$ we have $$0  = \sum\limits_{s \in I} 2\lambda_s y_s  + \sum\limits_{t \in J(z)} 2\mu_t y_t,$$ that can be simplified to $$0  = \sum\limits_{s \in I} \lambda_s y_s  + \sum\limits_{t \in J(z)} \mu_t y_t,$$ as stated. In addition, the condition ${\partial \mathcal{L}}/{\partial \xi_i} = 0$ is transformed into $$ 0  = -2\lambda_i - \delta_i + C_+, \quad  i \in I:y_i=+1$$ and $$ 0  = -2\lambda_i - \delta_i + C_-, \quad  i \in I:y_i=-1.$$}

{Furthermore, since these results must be equivalent to the case if we had maintained the previously removed constraint, we have $\mu_t = 0$ when $z_t=0, \quad t \in J$ and $\mu_t \geq 0$ when $z_t=1, \quad t \in J$. This can be summarized as $0 \leq \mu_t \leq M_2z_t,\quad t \in J$. Also, as usual, $\delta_i$ is removed since we add $$0 \leq \lambda_i \leq C_+/2, \quad  i \in I:y_i=+1$$ and $$0 \leq \lambda_i \leq C_-/2, \quad  i \in I:y_i=-1,$$ as we know that $\delta_i \geq 0$. Therefore, the KKT conditions result:
\[
\begin{array}{llll}
{\omega} & = & \sum\limits_{s \in I} \lambda_s y_s {x}_s+ \sum\limits_{t \in J} \mu_t y_t {x}_t \\
0 & = & \sum\limits_{s \in I} \lambda_s y_s  + \sum\limits_{t \in J} \mu_t y_t \\
0 & \leq & \lambda_s \leq C_+/2  & s\in I:y_i=+1 \\
0 & \leq & \lambda_s \leq C_-/2  & s\in I:y_i=-1 \\
0 & \leq & \mu_t  \leq  {M_2} z_t& t \in J.
\end{array}
\]}
\noindent

{Note that we have replaced all the $J(z)$ by $J$ using the previous clarification.}

{Thus, substituting the previous expressions into the second optimization problem, the partial dual of such problem can be calculated, yielding}

\[
\begin{array}{lllll}
\min\limits_{{z}}   & & &  {\min\limits_{{\lambda},{\mu}, \beta, {\xi}} } & \left(  \sum\limits_{s \in I} \lambda_s y_s {x}_s+ \sum\limits_{t \in J} \mu_t y_t {x}_t \right)^\top
\left(  \sum\limits_{s \in I} \lambda_s y_s {x}_s+ \sum\limits_{t \in J} \mu_t y_t {x}_t \right)  +\\
& & & &  {+ C_+\sum\limits_{i \in I : y_i =+1} \xi_i  + C_-\sum\limits_{i \in I : y_i =-1} \xi_i } \\
\mbox{s.t.} & z_j \in \{0,1\} & j \in J  & {\mbox{s.t.}}  & {y_i \left( \left(\sum\limits_{s \in I} \lambda_s y_s {x}_s+ \sum\limits_{t \in J} \mu_t y_t {x}_t\right)^\top {x}_i  + \beta \right)
 \geq 1 - \xi_i} \quad {i \in I}\\
 & {\hat{p}_\ell \geq {p_{0}^*}_{\ell}}  & \ell \in L &  & { y_j\left( 	\left( \sum\limits_{s \in I} \lambda_s y_s {x}_s+ \sum\limits_{t \in J} \mu_t y_t {x}_t\right)^\top {x}_j +
 \beta \right) \geq 1 -{M_1}(1-z_j)} \quad {j \in J} \\
& &  &  & { \xi_i \geq 0} \quad {i \in I }\\
& & & & {\sum\limits_{i \in I} \lambda_i y_i  + \sum\limits_{j \in J} \mu_j y_j = 0}\\
& &  &  & {{ 0 \leq \lambda_i \leq C_+/2} \quad {i \in I:y_i=+1 }}\\
& &  &  & {{ 0 \leq \lambda_i \leq C_-/2} \quad {i \in I:y_i=-1 }}\\
& &  & &  { 0 \leq \mu_j \leq {M_2}z_j} \quad {j \in J}.
\end{array}
\]

Finally, since this problem only depends on the observation via the inner product, we can use the kernel trick and Problem~(CSVM) is obtained.

\bibliographystyle{spcustom}

\bibliography{CSVM_ADAC}

\end{document}